\pdfoutput=1
\documentclass[11pt]{article}

\usepackage[]{acl}

\usepackage{times}
\usepackage{latexsym}

\usepackage[T1]{fontenc}
\usepackage[utf8]{inputenc}

\usepackage{microtype}
\usepackage{graphicx}

\usepackage{multirow}
\usepackage{inconsolata}
\usepackage{booktabs}
\usepackage{amsmath}

\title{Extending LLMs' Context Window with 100 Samples}

\author{
    Yikai Zhang\textsuperscript{1,2,3}\quad 
    Junlong Li\textsuperscript{1,3}\quad 
    Pengfei Liu\textsuperscript{1,2,3}\thanks{\,\, Corresponding author}\\
    \textsuperscript{1} Shanghai Jiao Tong University \quad
    \textsuperscript{2} Shanghai Artificial Intelligence Lab \\
    \textsuperscript{3} Generative AI Research Lab (GAIR)}

\begin{document}
\maketitle

\begin{abstract}
Large Language Models (LLMs) are known to have limited extrapolation ability beyond their pre-trained context window, constraining their application in downstream tasks with lengthy inputs. Recent studies have sought to extend LLMs' context window by modifying rotary position embedding (RoPE), a popular position encoding method adopted by well-known LLMs such as LLaMA, PaLM, and GPT-NeoX. However, prior works like Position Interpolation (PI) and YaRN are resource-intensive and lack comparative experiments to assess their applicability. In this work, we identify the inherent need for LLMs' attention entropy (i.e. the information entropy of attention scores) to maintain stability and introduce a novel extension to RoPE which combines adjusting RoPE's base frequency and scaling the attention logits to help LLMs efficiently adapt to a larger context window. We validate the superiority of our method in both fine-tuning performance and robustness across different context window sizes on various context-demanding tasks. Notably, our method extends the context window of LLaMA-2-7B-Chat to 16,384 with only \textit{100} samples and 6 training steps, showcasing extraordinary efficiency. Finally, we also explore how data compositions and training curricula affect context window extension for specific downstream tasks, suggesting fine-tuning LLMs with lengthy conversations as a good starting point. We release our code and SFT data at \url{https://github.com/GAIR-NLP/Entropy-ABF}.
\end{abstract}

\section{Introduction}
\label{intro}
Large Language Models (LLMs) are typically pre-trained with a pre-defined context window size. For instance, LLaMA 2~\citep{touvron2023llama} is pre-trained on sequences of 4,096 tokens. When exceeding the pre-trained context window, the performance of LLMs tends to deteriorate primarily due to the limited length extrapolation ability of their position encoding methods~\citep{kazemnejad2023impact}. The limited context window affects LLMs' practicality for ever-increasing context-demanding tasks such as few-shot learning~\citep{brown2020language}, long document summarization~\citep{huang2021efficient} and repository-level code completion~\citep{liu2023repobench}. Consequently, there is an urgent need to extend LLMs' context window.

To meet this pressing demand, recent works have witnessed progress in context window extension in both fine-tuned and non-fine-tuned scenarios by extending Rotary Position Embedding (RoPE)~\citep{su2021roformer}, a widely-used position encoding method adopted by state-of-the-art LLMs such as LLaMA~\citep{touvron2023llama1, touvron2023llama}, PaLM~\citep{chowdhery2023palm, anil2023palm} and GPT-NeoX~\citep{black2022gpt}. For example, Position Interpolation (PI)~\citep{kaiokendev, chen2023extending} linearly down-scales the input tokens' position indices and achieves improved fine-tuning results. NTK-Aware scaling~\citep{blocntkaware} and adjusted base frequency (ABF)~\citep{xiong2023effective} modify the base frequency of RoPE, leading to enhanced results in fine-tuning and non-fine-tuning scenarios respectively. NTK-By-Parts scaling~\citep{blocntkparts} treats different dimensions differently and reports even better fine-tuning outcomes. More recently, YaRN~\citep{peng2023yarn} proposes scaling the attention logits given its beneficial effects on language modeling perplexity. They combine this technique with NTK-By-Parts scaling and report the best long-context performance among existing RoPE-extension methods.

However, the underlying rationale behind the efficiency of YaRN's scaling operation remains poorly understood. In this study, we provide an interpretation of this technique by analyzing its effect on stabilizing the information entropy of models' attention scores. Through our analysis, we introduce our own RoPE-extension method termed ``entropy-aware ABF'', which combines ABF with a sophisticated utilization of dynamic attention scalar.

Moreover, despite the \textit{individual-reported} efficacy of previous RoPE-extension methods, there's a lack of comprehensive \textit{comparative} analysis where different methods are put in the same evaluation testbed.
This study also addresses this gap by answering three key questions pertinent to context window extension in real-world applications: (1) Which method exhibits the best supervised fine-tuning performance on context-demanding downstream tasks? (2) How can each method efficiently utilize training data?  (3) Do models trained with these methods have a robust performance across varying context window sizes?

To answer the above questions, we conduct experiments on a diverse set of context-demanding tasks from LongBench~\citep{bai2023longbench}, manipulating the training data amounts and prompt lengths to evaluate fine-tuned models across different dimensions. The experiment results demonstrate that models trained with our method surpass all baselines in long-context fine-tuning performance and also maintain a robust performance across various context window sizes. Notably, with only \textit{100} long conversations from ShareGPT~\citep{vicuna2023} and 6 training steps, using four A100 GPUs for approximately 6 minutes, our method produces a model with competent performance across 12 selected context-demanding tasks. Finally, we explore the influence of data compositions and training curricula on context window extension for a given long context downstream task, suggesting fine-tuning the model on lengthy ShareGPT conversations as a good starting point.
\def\vf{\mathbf{f}}
\def\vh{\mathrm{h}}
\def\vx{\mathbf{x}}
\def\vk{\mathbf{k}}
\def\vq{\mathbf{q}}
\def\vu{\mathbf{u}}
\def\di{\mathrm{i}}

\section{Preliminaries}
\paragraph{Rotary Position Embedding (RoPE)}
Given a position index $m \in [1, c]$ and an embedding vector $\vx:= [x_0, x_1, \ldots, x_{d-1}]^\top$, where $d$ is the dimension of each attention head, RoPE considers each pair of elements along the feature dimension of the embedding vector as complex numbers and encodes position information by rotating them. The vector-valued complex function $\vf(\vx, m)$ defined by RoPE is as follows:
\begin{equation}
\resizebox{0.65\linewidth}{!}{$
    \vf(\vx,m) = 
    \begin{bmatrix}
        (x_0 + \di x_1) e^{\di m \theta_1},\\
        (x_2 + \di x_3) e^{\di m \theta_2},\\
        \ldots, \\
        (x_{d-2} + \di x_{d-1}) e^{\di m \theta_{d/2}}
    \end{bmatrix}$}
\label{eq: rope}  
\end{equation}
$\di := \sqrt{-1}$ is the imaginary unit and $\theta_j = \text{b}^{-2j/d}$, where $b$ denotes the base frequency of RoPE and is set to $10,000$ by default.

In application, RoPE is applied to both query and key embeddings through the following equation:
\begin{equation}
\resizebox{0.89\linewidth}{!}{$
    \vf(\vx,m) = 
    \begin{bmatrix}
    	x_0\\
    	x_1\\
    	x_2\\
    	x_3\\
    	\vdots\\
    	x_{d-2}\\
    	x_{d-1}
    \end{bmatrix}
    \otimes
    \begin{bmatrix}
    	\cos(m\theta_0) \\
    	\cos(m\theta_0) \\
    	\cos(m\theta_1) \\
    	\cos(m\theta_1) \\
    	\vdots \\
    	\cos(m\theta_{(d-1)/2}) \\
    	\cos(m\theta_{(d-1)/2}) 
    \end{bmatrix}
    +
    \begin{bmatrix}
    	-x_1\\
    	x_0\\
    	-x_3\\
    	x_2\\
    	\vdots\\
    	-x_{d-1}\\
    	x_{d-2}
    \end{bmatrix}
    \otimes
    \begin{bmatrix}
    	\sin(m\theta_0)\\
    	\sin(m\theta_0)\\
    	\sin(m\theta_1)\\
    	\sin(m\theta_1)\\
    	\vdots\\
    	\sin(m\theta_{{(d-1)}/2})\\
    	\sin(m\theta_{{(d-1)}/2})
    \end{bmatrix}$}
\end{equation}

The fundamental components of RoPE are a series of trigonometric coefficients, each encoding position information of different frequencies.

We represent these trigonometric coefficients with the following function to uniquely identify RoPE and its variants:
\begin{equation}
    h(m, b, t) = \sqrt{t} * cos(\frac{m}{b^{\frac{2j}{d}}})\text{ or } \sqrt{t} * sin(\frac{m}{b^{\frac{2j}{d}}})
    \label{eq: repre}
\end{equation}
where $m$ is the position index of the query token, $b$ is the base frequency for RoPE, and $t$ is the scaling factor for attention logits. Note that $\sqrt{t}$ is used in the equation because RoPE rotates process both the query and key embeddings.

Before introducing RoPE-extension methods that enable better context window extension, we define context scaling factor $s = \frac{c'}{c}$, which is the ratio between the target context window $c'$ and the pre-trained context window $c$. It is of special use to those methods that extend RoPE according to a given target context window size.

\paragraph{Position Interpolation (PI)} PI~\citep{chen2023extending, kaiokendev} linearly interpolates the input position index $m$ to $\frac{m}{s}$ so that it falls within the original context window size. \citet{chen2023extending} demonstrate that direct fine-tuning of LLaMA~\citep{touvron2023llama1} with an extended context window results in minimal improvement, as the effective context window of the model only increases from 2,048 to 2560 after 10,000 training steps on sequences of length 8,192. By contrast, PI succeeds in extending the context window of LLaMA to 32,768 with only 1,000 training steps.

\paragraph{NTK-Aware} NTK-Aware scaling~\citep{blocntkaware} hypothesize that interpolating all dimensions equally, as done by PI, may result in loss of high-frequency information. Therefore, NTK-Aware scaling introduces a nonlinear interpolation strategy by adjusting the base frequency $b$ of RoPE to $b^{\frac{d}{d-2}}$. This modification scales the low-frequency components of RoPE to a similar extent as PI, while only slightly altering the high-frequency components to avoid disturbing high-frequency information. NTK-Aware extends models' context window size without training. However, this method can't benefit as much as PI from additional training on longer sequences as suggested by~\citep{peng2023yarn}.

\paragraph{NTK-By-Parts} NTK-By-Parts~\citep{blocntkparts} holds that stretching all the RoPE components either by a scaling factor $s$ or a base transformation results in token embeddings being closer to each other, impeding LLMs from effectively capturing local relationships between adjacent tokens. To address this issue, NTK-By-Parts scales $\theta(j)$ by a factor $\frac{1-\gamma(j)}{s}+\gamma(j)$, with $\gamma(j)$ being assigned 0 for high frequencies, 1 for low frequencies, and a predetermined constant within the range of 0 to 1 for intermediate frequencies. According to~\citep{peng2023yarn}, this method performs better than PI and NTK-Aware scaling for both fine-tuned and non-fine-tuned models.

\paragraph{YaRN} Yarn~\citep{peng2023yarn} empirically observes that introducing a temperature $t$ to scale the attention logits before the softmax function improves models' language modeling performance. They find the optimal value of $\sqrt{t}=0.1\ln{s}+1$ by fitting the lowest perplexity curve against various context scaling factors $s$. They combine their finding with NTK-By-Parts scaling and term this method YaRN (Yet another RoPE extensioN method). YaRN reports the best long-context performance on language modeling tasks among existing methods.

\paragraph{Adjusted Base Frequency (ABF)} ABF~\citep{xiong2023effective} simply changes the base frequency of RoPE to 50,000. Both theoretical analysis and experiment are conducted to validate the efficacy of this method. \citet{xiong2023effective} proves that ABF minimizes the distance of its embedded vectors from the ones using the original RoPE, which helps leverage the pre-training results. They empirically validate the efficacy of ABF by showing a lower perplexity on language modeling tasks and a longer effective context window in the first-sentence-retrieval task.

Table~\ref{tab: RoPE} highlights the difference between RoPE and its variants by specifying the different $m$, $b$, and $t$ they use in Equation~\ref{eq: repre} and whether they require additional training for context window extension:
\begin{table}[!htbp]
    \centering
    \resizebox{0.5\textwidth}{!}{
        \begin{tabular}{lcccc}
        \toprule
        \textbf{Method} & \textbf{$m$} & \textbf{$b$} & $t$ & \textbf{Additional Training} \\ \midrule
        RoPE & $m$ & $10,000$ & $1$ & -\\
        PI & $m/s$ & $10,000$ & $1$ & continual pre-train \\
        NTK-Aware & $m$ & $10,000^{\frac{d-2}{d}}$ & $1$ & - \\
        NTK-By-Parts & $(\frac{1-\gamma(j)}{s}+\gamma(j))m$ & $10,000$ & $1$ & continual pre-train\\
        YaRN & $(\frac{1-\gamma(j)}{s}+\gamma(j))m$ & $10,000$ & $0.1ln(s) + 1$ & continual pre-train\\
        ABF & $m$ & $500,000$ & $1$ & continual pre-train\\
        \bottomrule
        \end{tabular}
    }
    \vspace{3pt}
    \caption{An overview of Rotary Position Embedding (RoPE) and its variants represented by Equation~\ref{eq: repre}.}
    \label{tab: RoPE}
\end{table}
\section{Proposal Method}
YaRN~\citep{peng2023yarn} introduces a scaling factor $t$ on the attention logits based on empirical evidence indicating its beneficial effects on language modeling perplexities. However, the underlying rationale behind this technique remains poorly understood. In this section, we first introduce an interpretation of this technique, which motivates our method.

\subsection{Interpretation of YaRN's Scaling Factor}
In Transformer models' attention mechanism~\citep{vaswani2017attention}, the Softmax function forces attention scores assigned to contextual tokens to sum to one while concurrently preventing any individual score from becoming zero. Consequently, with an increasing number of input tokens, LLMs will theoretically distribute more attention across more tokens and lead to a rise in what we refer to as ``attention entropy'', which quantifies the randomness within the distribution of attention scores and is calculated using the following equation:
\begin{equation}
    \text{attention\_entropy}= \sum_{i}p_i\ln{p_i}
\label{eq: entropy}
\end{equation}
where $p_i$ is the attention scores assigned to contextual tokens.

\begin{figure}[htbp]
    \centering
    \includegraphics[width= 1\linewidth]{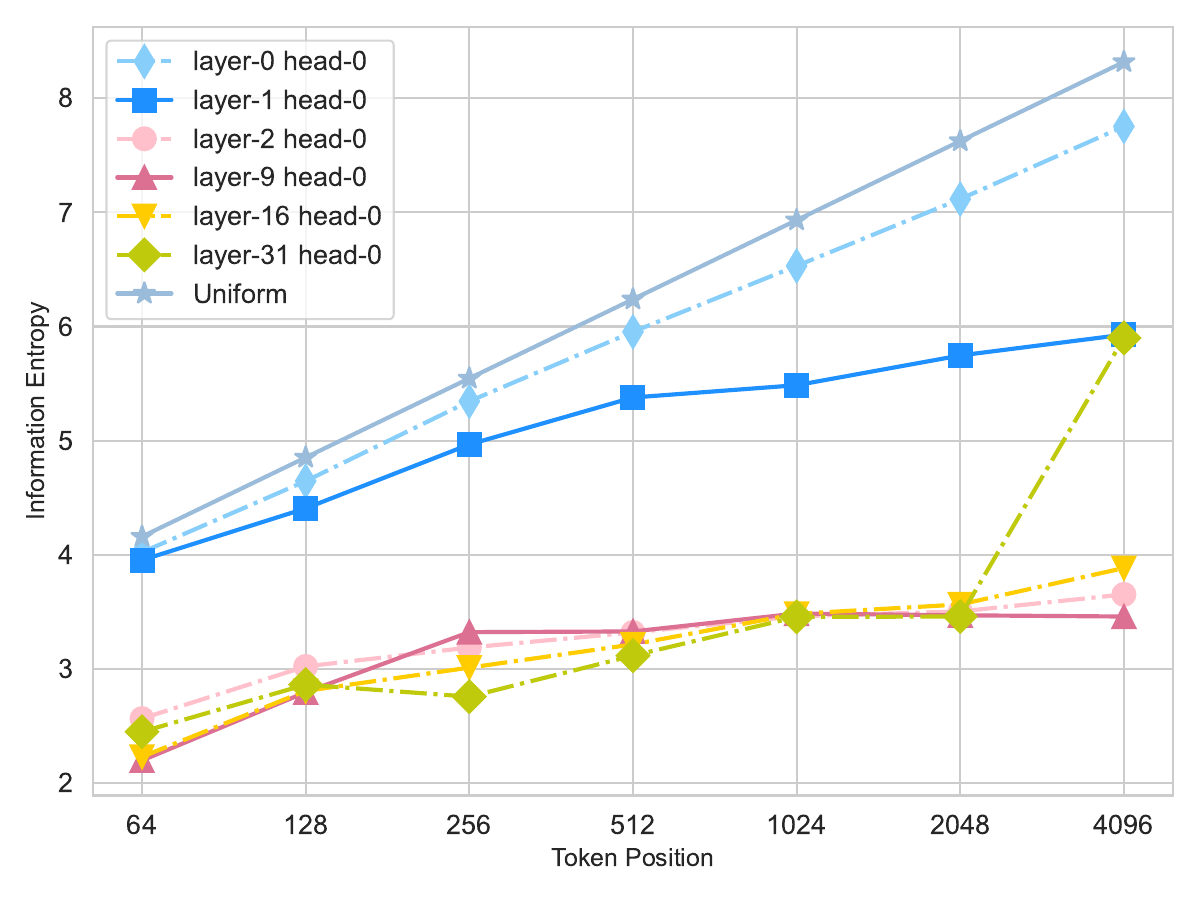}
    \caption{Visualization of the averaged attention entropy for query tokens at different input positions in the LLaMA-2-7B-chat model across the selected 128 documents from the Pile-arXiv dataset~\citep{gao2020pile}. ``Uniform'' represents a uniform attention score distribution, which corresponds to $\text{attention\_entropy}=\ln{n}$ with $n$ denoting the number of contextual tokens.}
    \label{fig: attn_observation}
\end{figure}

To validate the aforementioned theoretical effect, we utilized LLaMA-2-7B-Chat~\citep{touvron2023llama} to process 128 randomly chosen documents from the Pile dataset~\citep{gao2020pile}. 
We collect the attention scores assigned to contextual tokens for query tokens at different input positions to simulate varying numbers of contextual tokens. Subsequently, we compute the information entropy for these attention scores on different model layers via Equation~\ref{eq: entropy}. The resulting average attention entropies over our randomly sampled documents are visualized in Figure~\ref{fig: attn_observation}.

Counterintuitively, only the first two model layers demonstrate a steady rise in attention entropy.
Interestingly, we even observe that the attention entropies of all the subsequent layers remain remarkably similar when the number of contextual tokens increases from 1,024 to 2,048.

This finding of LLMs maintaining a stable attention entropy in subsequent model layers when contextual tokens are directly doubled leads us to posit that possessing a certain degree of length-invariance in attention entropy in these layers is an important inherent characteristic of LLMs for functioning properly. When modeling longer sequences than encountered in the pretraining stage, LLMs might fail to concentrate well, leading to a performance drop. Thanks to the exponential function in Softmax, scaling the attention logits reduces attention entropy, thereby explaining why it leads to improvements in language modeling tasks when modeling lengthy inputs as observed in YaRN~\citep{peng2023yarn}.

\subsection{Design Principles}
\label{design}
Previous works have explored different scaling factors on the attention logits with different motivations.
\citet{chiang2022overcoming} scales the attention logits by $\log n$, with $n$ representing the length of the longest training sequence, to enhance the model's extrapolation ability in downstream tasks such as machine translation.

More recently, YaRN~\citep{peng2023yarn} introduces the scaling factor $t = 0.1\ln s+1$ by fitting the lowest perplexity curve in language modeling tasks. They combine these scaling factors with NTK-By-Parts scaling and observe improved fine-tuning long-context performance on language modeling tasks.

ReRoPE~\citep{rerope2023} utilized a dynamic scaling factor that takes into account the number of contextual tokens for each input position:
$t = \log_c{m}$, where $c$ denotes the pre-trained context window size and $m$ represents the position index of input tokens. By introducing this scaling factor during the pre-training stage, ReRoPE demonstrates enhanced extrapolation ability in language modeling tasks, which is also observed in YaRN.

We propose ``entropy-aware ABF'' with the following design principles:

\noindent \textbf{(1). Dynamic Attention Scaling}: 
Both PI and YaRN employ a \textit{constant} scaling factor for all input positions, which may excessively stretch the attention logits at the front positions and hinder the model's ability to extrapolate to longer sequences.
Instead of using a constant scaling factor, we propose using a dynamic factor like ReRoPE that takes into account the number of contextual tokens for each input position. This allows the model to adjust the attention weights more flexibly based on the level of randomness in the distribution of attention scores.

\noindent \textbf{(2). Layer-dependent}: 
All the existing works apply the scalar indiscriminately to all model layers. However, based on our observations in Figure~\ref{fig: attn_observation} that the first two layers consistently exhibit a near-uniform attention pattern and only the latter layers demonstrate the tendency to maintain concentration, we propose not to intervene in the first two layers to align with the model's inherent characteristics.

\noindent \textbf{(3). Facilitate Context Window Extension}: Furthermore, we hypothesize that learning to maintain concentration when processing lengthy sequences is critical to context window extension, and scaling the attention logits can serve as an inductive bias that facilitates this process. This motivates us to combine ``scaling the attention logits'' with ABF during the supervised fine-tuning stage. To leverage the pretraining results, we also propose the avoidance of modifying the attention logits within the pre-trained context window by setting a lower bound to $t$.

Our ultimate scaling factor $t$ is depicted below:
\begin{equation*}
t = \begin{cases}
    1, & \text{if layer index is 0 or 1}\\
    \max(\log_c{i}, 1), & \text{o.w.}\\
\end{cases}
\end{equation*}
\section{Experiments}
To analyze the real-world applicability of different RoPE-extension methods, we test the long-context performance of models trained with these methods on selected tasks from LongBench~\citep{bai2023longbench} and answer the three research questions we propose in Section~\ref{intro} by adjusting training data amount and context window sizes. Finally, we also explore efficient data compositions and training curricula on context window extension for given downstream tasks.

\begin{table*}[!t]
    \centering
        \caption{Experiment results on selected tasks from LongBench. Model names with a trailing asteroid are reported from the LongBench paper. We name our trained models after their RoPE-extension methods.}
    \resizebox{\textwidth}{!}{
        \begin{tabular}{lccccccccccccc}
        \toprule
        \multirow{2}{*}{\textbf{\Large Model}} & \multicolumn{3}{c}{\textbf{\Large Singl-Doc QA}} & \multicolumn{3}{c}{\textbf{\Large Multi-Doc QA}} & \multicolumn{3}{c}{\textbf{\Large Summarization}} & \multicolumn{3}{c}{\textbf{\Large Few-shot Learning}} & \multirow{2}{*}{\textbf{\Large Macro}} \\
        \cmidrule(lr){2-4} \cmidrule(lr){5-7} \cmidrule(lr){8-10} \cmidrule(lr){11-13} 
        & \textbf{\Large NQA} & \textbf{\Large QAPR} & \textbf{\Large MFQA\_en} & \textbf{\Large HPQA} & \textbf{\Large WMQA} & \textbf{\Large MSQ} & \textbf{\Large GR} & \textbf{\Large QMSM} & \textbf{\Large MNWS} & \textbf{\Large TREC} & \textbf{\Large TRVQA} & \textbf{\Large SMSM} \\ \midrule
        \Large Llama2-7B-chat-4k* & \Large 18.7 & \Large 19.2 & \Large 36.8 & \Large 25.4 & \Large 32.8 &\Large 9.4 & \Large 27.3 & \Large 20.8 & \Large 25.8 &\Large 61.5 &\Large 77.8 &\Large 40.7 & \Large 33.0\\
        \Large LongChat-v1.5-7B-32k* & \Large 16.9 & \Large 27.7 & \Large 41.4 & \Large 31.5 & \Large 20.6 & \Large 9.7 & \Large 30.8 & \Large 22.7 & \Large 26.4 & \Large 63.5 & \Large 82.3 & \Large 34.2 & \Large 34.0 \\
        \Large Vicuna-v1.5-7B-16k* & \Large 19.4 & \Large 26.1 & \Large 38.5 & \Large 25.3 & \Large 20.8 & \Large 9.8 & \Large 27.9 & \Large 22.8 & \Large 27.2 & \Large 71.5 & \Large 86.2 & \Large 40.8 & \Large 34.7 \\
        \Large PI & \Large 20.1 & \Large 30.4 & \Large 45.3 & \Large 26.1 & \Large 30.1 & \Large 9.9 & \Large 28.1 & \Large 23.7 & \Large 26.6 & \Large 68.0 & \Large 84.9 & \Large 42.5 & \Large 36.3 \\
        \Large NTK-By-Parts & \Large 15.9 & \Large 31.1 & \Large 40.1 & \Large 25.4 & \Large 26.6 & \Large 7.2 & \Large 26.7 & \Large 22.4 & \Large 26.9 & \Large 68.5 & \Large 82.8 & \Large 42.9 & \Large 34.7 \\
        \Large Yarn & \Large 20.3 & \Large 28.9 & \Large 42.8 & \Large 27.8 & \Large 30.7 & \Large 7.2 & \Large 27.4 & \Large 22.5 & \Large 26.8 & \Large 66.0 & \Large 85.6 & \Large 42.6 & \Large 35.7 \\
        \Large ABF & \Large 24.6 & \Large 32.8 & \Large 45.6 & \Large 35.1 & \Large 30.3 & \Large 15.2 & \Large 30.8 & \Large 23.0 & \Large 27.4 & \Large 71.0 & \Large 84.7 & \Large 42.7 & \Large 38.6 \\
        \Large Ours & \Large 21.9 & \Large 31.0 & \Large 47.1 & \Large 40.1 & \Large 32.7 & \Large 15.1 & \Large 32.3 & \Large 23.0 & \Large 27.1 & \Large 70.5 & \Large 86.7 & \Large 42.0 & \Large \textbf{39.1} \\
        \bottomrule
        \end{tabular}
    }
    \label{tab: longbench}
\end{table*}

\subsection{General Setup}
\paragraph{Model Variants} We use LLaMA-2-7B-Chat~\citep{touvron2023llama} given its popularity. We only modify RoPE while leaving the model architecture unchanged.

\paragraph{Training} Previous works~\citep{chen2023extending, xiong2023effective, peng2023yarn} adopt a similar training curriculum by first continual pre-training the LLaMA base model to adapt to the modified position embeddings and then fine-tune on target long-context downstream tasks. In contrast, we propose directly supervised fine-tuning of the Chat Model to evaluate the practical applicability of different RoPE-extension methods. We extend the context window of LLaMA-2-7B-Chat to 16k with detailed training setups available in Appendix~\ref{appdix: training}.

\paragraph{SFT Data} We curate a dataset of 3.5k lengthy conversations from ShareGPT\footnote{https://huggingface.co/datasets/philschmid/sharegpt-raw}~\citep{vicuna2023}. Following the data cleaning pipeline in~\citep{zheng2023judging}, we kept English conversations only, excluded those with less than 10,000 tokens, and split longer conversations so that we have a maximum sequence length of 16,384 tokens.

\paragraph{Evaluation} Existing works primarily assess the efficacy of RoPE-extension methods through the examination of continual pre-trained models across language modeling tasks and synthetic tasks. For example, YaRN~\citep{chen2023extending} evaluates the perplexity scores and model performance on the passkey-retrieval task~\citep{mohtashami2023landmark} to quantify models' long-context performance. However, synthetic tasks like passkey retrieval deviate largely from real-world scenarios while language modeling tasks have also proved a rudimentary metric incapable of promising success in downstream tasks as suggested by~\citep{pal2023giraffe, Sun_Krishna_Mattarella-Micke_Iyyer_2021}.
In this work, we analyzed the long context performance of models with extended context windows on selected tasks from LongBench~\citep{bai2023longbench}. Our evaluation includes 12 tasks from four categories: single-document QA, multi-document QA, summarization, and few-shot learning to ensure a comprehensive evaluation of models' long-context capabilities. We intentionally exclude synthetic tasks and code completion tasks from LongBench because synthetic tasks deviate largely from real-world scenarios, and code completion tasks have performance conflicts with general instruction following abilities learned from ShareGPT conversations, as suggested by~\citep{dong2023abilities}.

\subsection{Measuring Long-Context Performance}
To answer the research question ``(1) Which method exhibits the best supervised fine-tuning performance on context-demanding downstream tasks?'', we fine-tune LLaMA-7B-Chat on 3.5k lengthy conversations and evaluate their long-context performance on LongBench.

Table~\ref{tab: longbench} illustrates the performance of each method, with some results reported from the LongBench paper~\citep{bai2023longbench}. We highlight our major observations here:

\textbf{1) Fine-tuning the models on lengthy conversation data is efficient for context window extension.} Both LongChat-v1.5-7B-32k and Vicuna-v1.5-7B-16k are open-source long-context models extended with PI~\citep{chen2023extending} through fine-tuning on large amounts of conversation data. For example, LongChat-v1.5-7B-32 is finetuned on 80k conversations. By fine-tuning the model on lengthy conversations only, our replicated PI-based model outperformed the open-source versions, confirming the efficacy of fine-tuning the model on lengthy conversations.

\textbf{2) PI yields better long-context fine-tuning results than YaRN.} While NTK-By-Parts and YaRN have lower perplexity in language modeling tasks, PI has better fine-tuning performance on long-context downstream tasks that are more related to practical scenarios. This finding corroborates the conclusion by~\citep{pal2023giraffe, Sun_Krishna_Mattarella-Micke_Iyyer_2021} that language modeling perplexity is a rudimentary metric incapable of promising success in downstream tasks. We hypothesize that while YaRN's scalar is efficient for language modeling tasks, its constant nature might affect model performance on downstream tasks.

\textbf{3) ABF-based models surpass the other methods by a significant margin.} Both ABF and our methods exhibit consistently superior fine-tuning performance on all 12 long-context tasks, demonstrating the efficacy of adjusting RoPE's base frequency to a large number (e.g. 50,000).

\subsection{Measuring Data Efficiency}
\begin{figure}[h]
    \centering
    \includegraphics[width=1\linewidth]{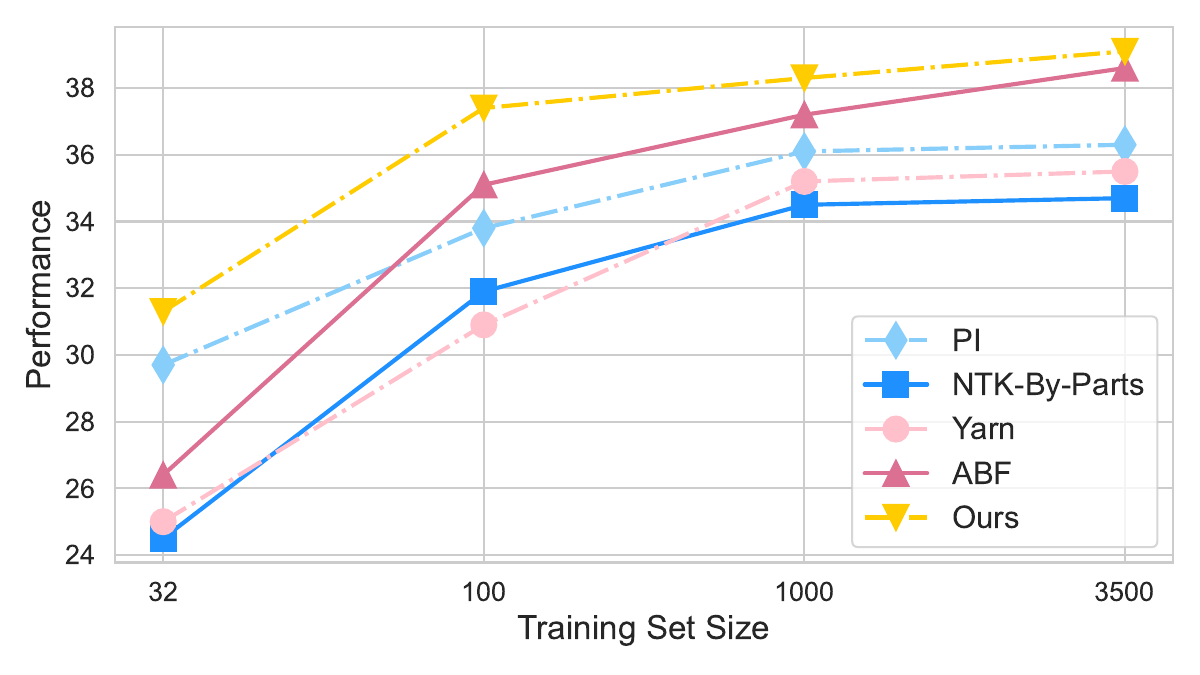}
    \caption{Long-Context Performance of RoPE-extending Methods with Different Amounts of Training Data}
    \label{fig: data}
\end{figure}

Data efficiency is an essential characteristic of RoPE-extension methods in context window extension practice, given both the sparsity of long training data and the high cost of training on long sequences. In this section, we explore the research question ``(2) How can each method efficiently utilize training data?'' by training the model respectively on 32, 100, 1k, and 3.5k conversations. The results are plotted in Figure~\ref{fig: data}, and the detailed results for each task are in Table~\ref{tab: data}.

We highlight our major observations below:

\textbf{1) ABF-based methods consistently benefit from increasing training data. } While all RoPE-extension methods exhibit improved performance with increased training data, the performance gain appears marginal for PI, NTK-By-Parts, and Yarn when the data amount increases from 1K to 3.5K. Only ABF-based methods consistently demonstrate performance gains.

\textbf{2) Entropy-Aware ABF demonstrates extraordinary data efficiency.} Notably, with a mere \textbf{100} training samples and \textbf{6} training steps, our method achieves competitive long-context performance that only lags marginally behind the ABF method trained on 3.5K samples. Without considering the cost of finetuning on downstream tasks, PI~\citep{chen2023extending} continue pre-trains LLaMA-7B~\citep{touvron2023llama1} for 1,000 steps with 64 batch size, YaRN~\citep{peng2023yarn} adopts 250 continual pre-training steps with the same batch size. Open source practice like Longchat~\citep{longchat2023} utilizes 80k conversations from ShareGPT for instruction tuning. Our work demonstrates the remarkable efficiency of entropy-aware ABF in context window extension, requiring less than 2\% of the training resources utilized by existing methodologies.

We also observe that the performance gap from ABF to our method is diminishing with the increase in training data. This phenomenon aligns with our hypothesis in Section~\ref{design} that while the ability to maintain concentration across lengthy inputs can be learned from training on more data, our method serves as an inductive bias that facilitates the learning process.

\subsection{Measuring Robustness across Context Windows}
A desirable attribute for RoPE-extension methods, when applied in practical context window extension settings, is that the models fine-tuned using these methods should maintain their performance on the original context window, while also demonstrating a certain degree of extrapolation capability beyond the fine-tuned length. 

To answer the research question ``(3) Do models trained with these methods have a robust performance across varying context window sizes?'', we follow LongBench~\citep{bai2023longbench} to assess the models across different context window sizes by truncating the prompt from the middle when the task length exceeds a designated context window size.

\begin{figure}[h]
    \centering
    \includegraphics[width=1\linewidth]{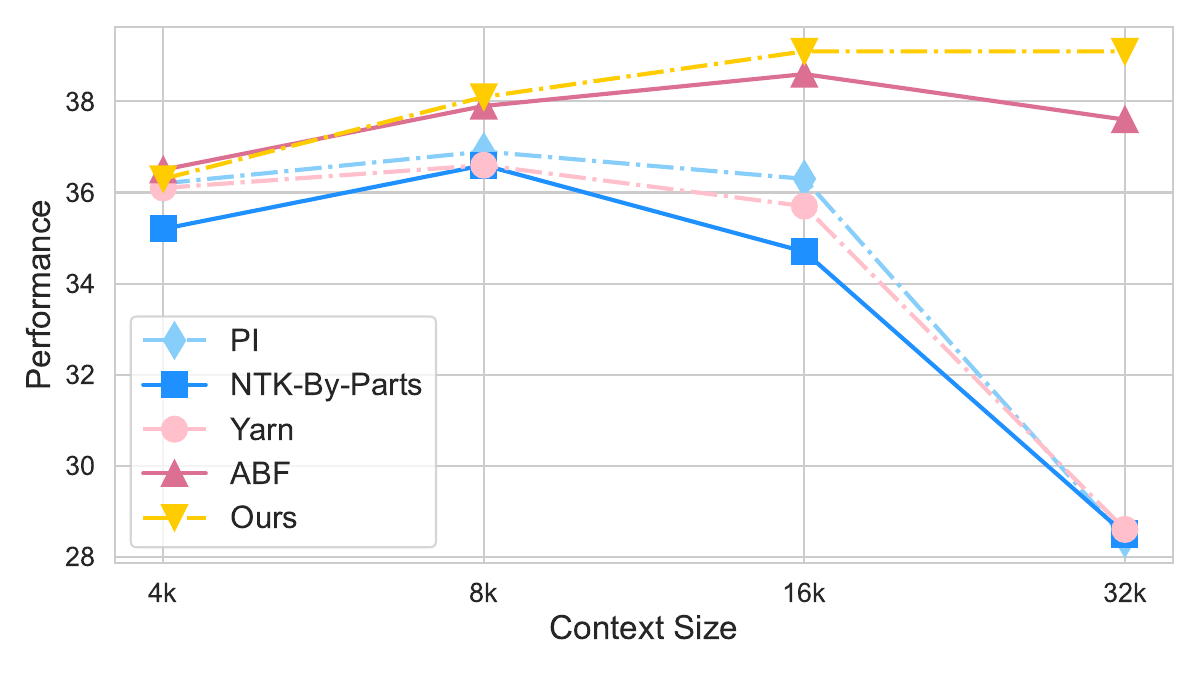}
    \caption{Long-Context Performance of RoPE-extending Methods with Different Context Window Sizes}
    \label{fig: ctx}
\end{figure}

The results are depicted in Figure~\ref{fig: ctx}. While there appears a performance gain for PI, NTK-By-Parts, and Yarn when the context size is enlarged from 4k to 8k, their performance degrades when the context is further enlarged to 16k, demonstrating their inability to leverage the full fine-tuning context window. In contrast, ABF and our proposed method consistently gain from a larger context window within fine-tuning length. Furthermore, entropy-aware ABF is the only method that can maintain the performance when directly extrapolating to 32k.

\subsection{Exploring the Optimal Training Data and Curriculums}
In this section, we explore efficient training data and curriculums for context window extension on given tasks.
An important consideration in practice is whether long in-domain training samples are indispensable for achieving success in context window extension for a particular downstream task. Specifically, we inquire whether short in-domain training samples only can still yield benefits in scenarios where lengthier samples are absent, which is often the case.
To answer the above questions, we conduct experiments with various training curriculums on GovReport~\citep{huang2021efficient} which is a widely used long context summarization task, and Longchat-Line-Retrieval~\citep{longchat2023}, a synthetic retrieval task. 

We evaluate both long (more than 8,092 tokens) and short tasks (within 4,096 tokens) to guarantee models' performance within the original context window while evaluating their long-context performance. When the training data is in-domain samples, we train the model for 4 epochs with a batch size of 8 and evaluate with the best epoch on the validation set. When the training data is 1,000 ShareGPT conversations, the model is trained for two epochs with a batch size of 32 and evaluated on the second epoch.

The results are displayed in Table~\ref{tab: datacomb}. We conclude that training the model on short in-domain samples produces suboptimal results, but starting from the model finetuned on 1,000 ShareGPT conversations yields comparable results to those finetuned on long in-domain samples, which suggests a good starting point for context window extension in practice. 

It might be strange that the line-retrieval task shows extremely poor performance when finetuned from the Chat model on long samples. We attribute it to the insufficient training of our method because the answer to the line retrieval task is short, and we only calculate losses on the model response tokens during the instruction tuning.

\begin{table}[!htbp]
    \centering
    \resizebox{0.5\textwidth}{!}{
        \begin{tabular}{lccccc}
        \toprule
        \textbf{Initialization} & \textbf{training data} & \textbf{GR-S} & \textbf{GR-L} & \textbf{LR-S} & \textbf{LR-L}
        \\ \midrule
        LLaMA 2 Chat & None & 30.84 & 0 & 76 & 0 \\
        LLaMA 2 Chat & Short & 37.91 & 33.6 & 74 & 26 \\
        LLaMA 2 Chat & Long & 38.24 & 36.45 & 10 & 2 \\
        Share1k & None & 34.10 & 31.14 & 88	& 48 \\
        Share1k & Short & 38.31	& 35.12	& 86 & 64 \\
        Share1k & Long & 38.93 & 35.56 & 92 & 66 \\
        Short & Share1k & 39.74	& 32.12	& 90 & 54 \\
        \bottomrule
        \end{tabular}
    }
    \vspace{3pt}
    \caption{Performance on two downstream tasks with different training curriculums. GR-S: GovReport-Short. GR-L: GovReport-Long. LR-S: Line Retrieval-Short. LR-L: LineRetrieval-Long. In the first column, \textbf{Share1k} means the fine-tuned result of the 7B Chat model on 1,000 ShareGPT conversations. \textbf{Short} means the fine-tuned result of the 7B chat model on short in-domain samples. In the second column, \textbf{None} means the model is directly tested. \textbf{Short} means short in-domain samples. \textbf{Long} means long in-domain samples.}
    \label{tab: datacomb}
\end{table}

\section{Related Work}
Extensive research has been done to enhance the long-context capacity of transformer models~\citep{vaswani2017attention} by overcoming two prominent obstacles: the quadratic time and space complexity of the attention mechanism~\citep{vaswani2017attention} and the inability of position encodings to generalize beyond the pre-trained context window.

\paragraph{More Efficient Transformers}
The vanilla attention mechanism in the Transformer architecture is known for its quadratic time and space complexity, which poses significant resource demands for transformer models when processing lengthy inputs. Various works have focused on conquering the complexity issue and proposing more efficient Transformers. Sparse transformers~\citep{child2019generating, Ye_Guo_Gan_Qiu_Zhang_2019, kitaev2020reformer, beltagy2020longformer, ainslie2020etc, zaheer2020big, ding2023longnet} replace the original full attention mechanism with a sparsified version to make the computation more efficient. Linear transformers~\citep{wang2020linformer, katharopoulos2020transformers, choromanski2020rethinking}, rather than forcing the attention mechanism to attend to fewer tokens, propose an alternative approach by leveraging low-rank matrix multiplication or linear dot-product of kernel feature maps to approximate the original attention mechanism, achieving linear time complexity. Meanwhile, retrieval-augmented models~\citep{guu2020retrieval, lewis2020retrieval, wu2022memorizing, bulatov2023scaling, tworkowski2023focused} integrate retrieval with attention. During inference time, these models avoid directly modeling lengthy inputs by retrieving information from an external memory that stores previous key-value pairs. While prior research primarily focuses on reducing FLOPs, the bottleneck of transformer inference on modern computing hardware has shifted to the overhead from memory access (IO). Multi-query attention (MQA)\citep{shazeer2019fast} and grouped-query attention (GQA)\citep{ainslie2023gqa}, for instance, address the memory-bandwidth cost associated with loading the large "keys" and "values" tensors in the multi-head attention mechanism by proposing the use of fewer "key" and "value" heads. Notably, GQA is employed in LLaMA2~\cite{touvron2023llama}. Additionally, FlashAttention~\citep{dao2022flashattention, dao2023flashattention} introduces an IO-aware exact attention approach that utilizes tiling to reduce memory IOs.

\paragraph{Generalizable Position Encoding}
Due to the attention mechanism's parallel nature, transformer models require position encoding (PE) methods to facilitate the integration of position information. The original transformer employed sinusoidal position encoding, which constitutes an absolute PE and exhibits limited generalization capability. Subsequently, this approach was refined to a learnable version~\citep{gehring2017convolutional}, as embraced by language model architectures such as GPT-3~\citep{brown2020language}. However, this adaptation completely compromises the extrapolation ability of position encoding methods. The advent of relative PE~\citep{shaw2018self} theoretically supports infinite input lengths. Nevertheless, despite recent advancements in relative PEs, such as T5 relative PE~\citep{raffel2020exploring}, RoPE~\citep{su2021roformer}, xPOS~\citep{sun2022length}, and ALiBi~\citep{press2021train}, it has been demonstrated by~\citep{kazemnejad2023impact} that all these methods fail when extrapolating significantly beyond the pre-trained context window.

\section{Conclusions}
In summary, through interpreting LLMs' inherent need to maintain concentration when processing lengthy sequences, we propose entropy-aware ABF by combining ABF with a sophisticated applied scalar that scales the attention logits. Our proposed method effectively extends the context window of RoPE-based LLMs, addressing their limitations when confronted with context-demanding tasks at a minimal cost. We empirically show the superiority of our method in both fine-tuning results and robustness across different context window sizes on various context-demanding tasks. Importantly, our method exhibits extraordinary data efficiency compared to other methods, deriving a competent long-context model on LongBench with only 100 samples and 6 training steps, less than 2\% of the training resources utilized by previous works. Finally, we provide valuable insights into context window extension for specific downstream tasks, suggesting training on lengthy ShareGPT conversations as a good starting point.

\section*{Acknowledgments}
We want to thank Zhengbao Jiang for his participation in the initial discussions. We thank Fan Nie and Fan Zhou for their invaluable advice throughout the paper-writing process.

\bibliography{anthology,custom}

\begin{thebibliography}{57}
\expandafter\ifx\csname natexlab\endcsname\relax\def\natexlab#1{#1}\fi

\bibitem[{Ainslie et~al.(2023)Ainslie, Lee-Thorp, de~Jong, Zemlyanskiy, Lebr{\'o}n, and Sanghai}]{ainslie2023gqa}
Joshua Ainslie, James Lee-Thorp, Michiel de~Jong, Yury Zemlyanskiy, Federico Lebr{\'o}n, and Sumit Sanghai. 2023.
\newblock Gqa: Training generalized multi-query transformer models from multi-head checkpoints.
\newblock \emph{arXiv preprint arXiv:2305.13245}.

\bibitem[{Ainslie et~al.(2020)Ainslie, Ontanon, Alberti, Cvicek, Fisher, Pham, Ravula, Sanghai, Wang, and Yang}]{ainslie2020etc}
Joshua Ainslie, Santiago Ontanon, Chris Alberti, Vaclav Cvicek, Zachary Fisher, Philip Pham, Anirudh Ravula, Sumit Sanghai, Qifan Wang, and Li~Yang. 2020.
\newblock Etc: Encoding long and structured inputs in transformers.
\newblock \emph{arXiv preprint arXiv:2004.08483}.

\bibitem[{Anil et~al.(2023)Anil, Dai, Firat, Johnson, Lepikhin, Passos, Shakeri, Taropa, Bailey, Chen et~al.}]{anil2023palm}
Rohan Anil, Andrew~M Dai, Orhan Firat, Melvin Johnson, Dmitry Lepikhin, Alexandre Passos, Siamak Shakeri, Emanuel Taropa, Paige Bailey, Zhifeng Chen, et~al. 2023.
\newblock Palm 2 technical report.
\newblock \emph{arXiv preprint arXiv:2305.10403}.

\bibitem[{Bai et~al.(2023)Bai, Lv, Zhang, Lyu, Tang, Huang, Du, Liu, Zeng, Hou et~al.}]{bai2023longbench}
Yushi Bai, Xin Lv, Jiajie Zhang, Hongchang Lyu, Jiankai Tang, Zhidian Huang, Zhengxiao Du, Xiao Liu, Aohan Zeng, Lei Hou, et~al. 2023.
\newblock Longbench: A bilingual, multitask benchmark for long context understanding.
\newblock \emph{arXiv preprint arXiv:2308.14508}.

\bibitem[{Beltagy et~al.(2020)Beltagy, Peters, and Cohan}]{beltagy2020longformer}
Iz~Beltagy, Matthew~E Peters, and Arman Cohan. 2020.
\newblock Longformer: The long-document transformer.
\newblock \emph{arXiv preprint arXiv:2004.05150}.

\bibitem[{Black et~al.(2022)Black, Biderman, Hallahan, Anthony, Gao, Golding, He, Leahy, McDonell, Phang et~al.}]{black2022gpt}
Sid Black, Stella Biderman, Eric Hallahan, Quentin Anthony, Leo Gao, Laurence Golding, Horace He, Connor Leahy, Kyle McDonell, Jason Phang, et~al. 2022.
\newblock Gpt-neox-20b: An open-source autoregressive language model.
\newblock \emph{arXiv preprint arXiv:2204.06745}.

\bibitem[{bloc97(2023{\natexlab{a}})}]{blocntkparts}
bloc97. 2023{\natexlab{a}}.
\newblock \href {https://github.com/jquesnelle/scaled-rope/pull/1} {{Add NTK-Aware interpolation "by parts" correction}}.

\bibitem[{bloc97(2023{\natexlab{b}})}]{blocntkaware}
bloc97. 2023{\natexlab{b}}.
\newblock \href {https://www.reddit.com/r/LocalLLaMA/comments/14lz7j5/ntkaware_scaled_rope_allows_llama_models_to_have/} {{NTK-Aware Scaled RoPE allows LLaMA models to have extended (8k+) context size without any fine-tuning and minimal perplexity degradation.}}

\bibitem[{Brown et~al.(2020)Brown, Mann, Ryder, Subbiah, Kaplan, Dhariwal, Neelakantan, Shyam, Sastry, Askell et~al.}]{brown2020language}
Tom Brown, Benjamin Mann, Nick Ryder, Melanie Subbiah, Jared~D Kaplan, Prafulla Dhariwal, Arvind Neelakantan, Pranav Shyam, Girish Sastry, Amanda Askell, et~al. 2020.
\newblock Language models are few-shot learners.
\newblock \emph{Advances in neural information processing systems}, 33:1877--1901.

\bibitem[{Bulatov et~al.(2023)Bulatov, Kuratov, and Burtsev}]{bulatov2023scaling}
Aydar Bulatov, Yuri Kuratov, and Mikhail~S Burtsev. 2023.
\newblock Scaling transformer to 1m tokens and beyond with rmt.
\newblock \emph{arXiv preprint arXiv:2304.11062}.

\bibitem[{Chen et~al.(2023)Chen, Wong, Chen, and Tian}]{chen2023extending}
Shouyuan Chen, Sherman Wong, Liangjian Chen, and Yuandong Tian. 2023.
\newblock Extending context window of large language models via positional interpolation.
\newblock \emph{arXiv preprint arXiv:2306.15595}.

\bibitem[{Chen et~al.(2016)Chen, Xu, Zhang, and Guestrin}]{chen2016training}
Tianqi Chen, Bing Xu, Chiyuan Zhang, and Carlos Guestrin. 2016.
\newblock Training deep nets with sublinear memory cost.
\newblock \emph{arXiv preprint arXiv:1604.06174}.

\bibitem[{Chiang and Cholak(2022)}]{chiang2022overcoming}
David Chiang and Peter Cholak. 2022.
\newblock Overcoming a theoretical limitation of self-attention.
\newblock \emph{arXiv preprint arXiv:2202.12172}.

\bibitem[{Chiang et~al.(2023)Chiang, Li, Lin, Sheng, Wu, Zhang, Zheng, Zhuang, Zhuang, Gonzalez, Stoica, and Xing}]{vicuna2023}
Wei-Lin Chiang, Zhuohan Li, Zi~Lin, Ying Sheng, Zhanghao Wu, Hao Zhang, Lianmin Zheng, Siyuan Zhuang, Yonghao Zhuang, Joseph~E. Gonzalez, Ion Stoica, and Eric~P. Xing. 2023.
\newblock \href {https://lmsys.org/blog/2023-03-30-vicuna/} {Vicuna: An open-source chatbot impressing gpt-4 with 90\%* chatgpt quality}.

\bibitem[{Child et~al.(2019)Child, Gray, Radford, and Sutskever}]{child2019generating}
Rewon Child, Scott Gray, Alec Radford, and Ilya Sutskever. 2019.
\newblock Generating long sequences with sparse transformers.
\newblock \emph{arXiv preprint arXiv:1904.10509}.

\bibitem[{Choromanski et~al.(2020)Choromanski, Likhosherstov, Dohan, Song, Gane, Sarlos, Hawkins, Davis, Mohiuddin, Kaiser et~al.}]{choromanski2020rethinking}
Krzysztof Choromanski, Valerii Likhosherstov, David Dohan, Xingyou Song, Andreea Gane, Tamas Sarlos, Peter Hawkins, Jared Davis, Afroz Mohiuddin, Lukasz Kaiser, et~al. 2020.
\newblock Rethinking attention with performers.
\newblock \emph{arXiv preprint arXiv:2009.14794}.

\bibitem[{Chowdhery et~al.(2023)Chowdhery, Narang, Devlin, Bosma, Mishra, Roberts, Barham, Chung, Sutton, Gehrmann et~al.}]{chowdhery2023palm}
Aakanksha Chowdhery, Sharan Narang, Jacob Devlin, Maarten Bosma, Gaurav Mishra, Adam Roberts, Paul Barham, Hyung~Won Chung, Charles Sutton, Sebastian Gehrmann, et~al. 2023.
\newblock Palm: Scaling language modeling with pathways.
\newblock \emph{Journal of Machine Learning Research}, 24(240):1--113.

\bibitem[{Dao(2023)}]{dao2023flashattention}
Tri Dao. 2023.
\newblock Flashattention-2: Faster attention with better parallelism and work partitioning.
\newblock \emph{arXiv preprint arXiv:2307.08691}.

\bibitem[{Dao et~al.(2022)Dao, Fu, Ermon, Rudra, and R{\'e}}]{dao2022flashattention}
Tri Dao, Dan Fu, Stefano Ermon, Atri Rudra, and Christopher R{\'e}. 2022.
\newblock Flashattention: Fast and memory-efficient exact attention with io-awareness.
\newblock \emph{Advances in Neural Information Processing Systems}, 35:16344--16359.

\bibitem[{Ding et~al.(2023)Ding, Ma, Dong, Zhang, Huang, Wang, and Wei}]{ding2023longnet}
Jiayu Ding, Shuming Ma, Li~Dong, Xingxing Zhang, Shaohan Huang, Wenhui Wang, and Furu Wei. 2023.
\newblock Longnet: Scaling transformers to 1,000,000,000 tokens.
\newblock \emph{arXiv preprint arXiv:2307.02486}.

\bibitem[{Dong et~al.(2023)Dong, Yuan, Lu, Li, Xue, Liu, Wang, Yuan, Zhou, and Zhou}]{dong2023abilities}
Guanting Dong, Hongyi Yuan, Keming Lu, Chengpeng Li, Mingfeng Xue, Dayiheng Liu, Wei Wang, Zheng Yuan, Chang Zhou, and Jingren Zhou. 2023.
\newblock How abilities in large language models are affected by supervised fine-tuning data composition.
\newblock \emph{arXiv preprint arXiv:2310.05492}.

\bibitem[{Gao et~al.(2020)Gao, Biderman, Black, Golding, Hoppe, Foster, Phang, He, Thite, Nabeshima et~al.}]{gao2020pile}
Leo Gao, Stella Biderman, Sid Black, Laurence Golding, Travis Hoppe, Charles Foster, Jason Phang, Horace He, Anish Thite, Noa Nabeshima, et~al. 2020.
\newblock The pile: An 800gb dataset of diverse text for language modeling.
\newblock \emph{arXiv preprint arXiv:2101.00027}.

\bibitem[{Gehring et~al.(2017)Gehring, Auli, Grangier, Yarats, and Dauphin}]{gehring2017convolutional}
Jonas Gehring, Michael Auli, David Grangier, Denis Yarats, and Yann~N Dauphin. 2017.
\newblock Convolutional sequence to sequence learning.
\newblock In \emph{International conference on machine learning}, pages 1243--1252. PMLR.

\bibitem[{Guu et~al.(2020)Guu, Lee, Tung, Pasupat, and Chang}]{guu2020retrieval}
Kelvin Guu, Kenton Lee, Zora Tung, Panupong Pasupat, and Mingwei Chang. 2020.
\newblock Retrieval augmented language model pre-training.
\newblock In \emph{International conference on machine learning}, pages 3929--3938. PMLR.

\bibitem[{Huang et~al.(2021)Huang, Cao, Parulian, Ji, and Wang}]{huang2021efficient}
Luyang Huang, Shuyang Cao, Nikolaus Parulian, Heng Ji, and Lu~Wang. 2021.
\newblock Efficient attentions for long document summarization.
\newblock \emph{arXiv preprint arXiv:2104.02112}.

\bibitem[{kaiokendev(2023)}]{kaiokendev}
kaiokendev. 2023.
\newblock \href {https://kaiokendev.github.io/til#extending-context-to-8k} {{Things I'm learning while training superhot.}}

\bibitem[{Katharopoulos et~al.(2020)Katharopoulos, Vyas, Pappas, and Fleuret}]{katharopoulos2020transformers}
Angelos Katharopoulos, Apoorv Vyas, Nikolaos Pappas, and Fran{\c{c}}ois Fleuret. 2020.
\newblock Transformers are rnns: Fast autoregressive transformers with linear attention.
\newblock In \emph{International conference on machine learning}, pages 5156--5165. PMLR.

\bibitem[{Kazemnejad et~al.(2023)Kazemnejad, Padhi, Ramamurthy, Das, and Reddy}]{kazemnejad2023impact}
Amirhossein Kazemnejad, Inkit Padhi, Karthikeyan~Natesan Ramamurthy, Payel Das, and Siva Reddy. 2023.
\newblock The impact of positional encoding on length generalization in transformers.
\newblock \emph{arXiv preprint arXiv:2305.19466}.

\bibitem[{Kitaev et~al.(2020)Kitaev, Kaiser, and Levskaya}]{kitaev2020reformer}
Nikita Kitaev, {\L}ukasz Kaiser, and Anselm Levskaya. 2020.
\newblock Reformer: The efficient transformer.
\newblock \emph{arXiv preprint arXiv:2001.04451}.

\bibitem[{Lewis et~al.(2020)Lewis, Perez, Piktus, Petroni, Karpukhin, Goyal, K{\"u}ttler, Lewis, Yih, Rockt{\"a}schel et~al.}]{lewis2020retrieval}
Patrick Lewis, Ethan Perez, Aleksandra Piktus, Fabio Petroni, Vladimir Karpukhin, Naman Goyal, Heinrich K{\"u}ttler, Mike Lewis, Wen-tau Yih, Tim Rockt{\"a}schel, et~al. 2020.
\newblock Retrieval-augmented generation for knowledge-intensive nlp tasks.
\newblock \emph{Advances in Neural Information Processing Systems}, 33:9459--9474.

\bibitem[{Li* et~al.(2023)Li*, Shao*, Xie, Sheng, Zheng, Gonzalez, Stoica, Ma, and Zhang}]{longchat2023}
Dacheng Li*, Rulin Shao*, Anze Xie, Ying Sheng, Lianmin Zheng, Joseph~E. Gonzalez, Ion Stoica, Xuezhe Ma, and Hao Zhang. 2023.
\newblock \href {https://lmsys.org/blog/2023-06-29-longchat} {How long can open-source llms truly promise on context length?}

\bibitem[{Liu et~al.(2023)Liu, Xu, and McAuley}]{liu2023repobench}
Tianyang Liu, Canwen Xu, and Julian McAuley. 2023.
\newblock Repobench: Benchmarking repository-level code auto-completion systems.
\newblock \emph{arXiv preprint arXiv:2306.03091}.

\bibitem[{Loshchilov and Hutter(2017)}]{Loshchilov_Hutter_2017}
Ilya Loshchilov and Frank Hutter. 2017.
\newblock Decoupled weight decay regularization.
\newblock \emph{Learning,Learning}.

\bibitem[{Mohtashami and Jaggi(2023)}]{mohtashami2023landmark}
Amirkeivan Mohtashami and Martin Jaggi. 2023.
\newblock Landmark attention: Random-access infinite context length for transformers.
\newblock \emph{arXiv preprint arXiv:2305.16300}.

\bibitem[{Pal et~al.(2023)Pal, Karkhanis, Roberts, Dooley, Sundararajan, and Naidu}]{pal2023giraffe}
Arka Pal, Deep Karkhanis, Manley Roberts, Samuel Dooley, Arvind Sundararajan, and Siddartha Naidu. 2023.
\newblock Giraffe: Adventures in expanding context lengths in llms.
\newblock \emph{arXiv preprint arXiv:2308.10882}.

\bibitem[{Peng et~al.(2023)Peng, Quesnelle, Fan, and Shippole}]{peng2023yarn}
Bowen Peng, Jeffrey Quesnelle, Honglu Fan, and Enrico Shippole. 2023.
\newblock Yarn: Efficient context window extension of large language models.
\newblock \emph{arXiv preprint arXiv:2309.00071}.

\bibitem[{Press et~al.(2021)Press, Smith, and Lewis}]{press2021train}
Ofir Press, Noah~A Smith, and Mike Lewis. 2021.
\newblock Train short, test long: Attention with linear biases enables input length extrapolation.
\newblock \emph{arXiv preprint arXiv:2108.12409}.

\bibitem[{Raffel et~al.(2020)Raffel, Shazeer, Roberts, Lee, Narang, Matena, Zhou, Li, and Liu}]{raffel2020exploring}
Colin Raffel, Noam Shazeer, Adam Roberts, Katherine Lee, Sharan Narang, Michael Matena, Yanqi Zhou, Wei Li, and Peter~J Liu. 2020.
\newblock Exploring the limits of transfer learning with a unified text-to-text transformer.
\newblock \emph{The Journal of Machine Learning Research}, 21(1):5485--5551.

\bibitem[{Rajbhandari et~al.(2020)Rajbhandari, Rasley, Ruwase, and He}]{rajbhandari2020zero}
Samyam Rajbhandari, Jeff Rasley, Olatunji Ruwase, and Yuxiong He. 2020.
\newblock Zero: Memory optimizations toward training trillion parameter models.
\newblock In \emph{SC20: International Conference for High Performance Computing, Networking, Storage and Analysis}, pages 1--16. IEEE.

\bibitem[{Rasley et~al.(2020)Rasley, Rajbhandari, Ruwase, and He}]{rasley2020deepspeed}
Jeff Rasley, Samyam Rajbhandari, Olatunji Ruwase, and Yuxiong He. 2020.
\newblock Deepspeed: System optimizations enable training deep learning models with over 100 billion parameters.
\newblock In \emph{Proceedings of the 26th ACM SIGKDD International Conference on Knowledge Discovery \& Data Mining}, pages 3505--3506.

\bibitem[{Ren et~al.(2021)Ren, Rajbhandari, Aminabadi, Ruwase, Yang, Zhang, Li, and He}]{ren2021zero}
Jie Ren, Samyam Rajbhandari, Reza~Yazdani Aminabadi, Olatunji Ruwase, Shuangyan Yang, Minjia Zhang, Dong Li, and Yuxiong He. 2021.
\newblock $\{$ZeRO-Offload$\}$: Democratizing $\{$Billion-Scale$\}$ model training.
\newblock In \emph{2021 USENIX Annual Technical Conference (USENIX ATC 21)}, pages 551--564.

\bibitem[{Shaw et~al.(2018)Shaw, Uszkoreit, and Vaswani}]{shaw2018self}
Peter Shaw, Jakob Uszkoreit, and Ashish Vaswani. 2018.
\newblock Self-attention with relative position representations.
\newblock \emph{arXiv preprint arXiv:1803.02155}.

\bibitem[{Shazeer(2019)}]{shazeer2019fast}
Noam Shazeer. 2019.
\newblock Fast transformer decoding: One write-head is all you need.
\newblock \emph{arXiv preprint arXiv:1911.02150}.

\bibitem[{Su(2023)}]{rerope2023}
Jianlin Su. 2023.
\newblock Rectified rotary position embeddings.
\newblock \url{https://github.com/bojone/rerope}.

\bibitem[{Su et~al.(2021)Su, Lu, Pan, Murtadha, Wen, and Liu}]{su2021roformer}
Jianlin Su, Yu~Lu, Shengfeng Pan, Ahmed Murtadha, Bo~Wen, and Yunfeng Liu. 2021.
\newblock Roformer: Enhanced transformer with rotary position embedding.
\newblock \emph{arXiv preprint arXiv:2104.09864}.

\bibitem[{Sun et~al.(2021)Sun, Krishna, Mattarella-Micke, and Iyyer}]{Sun_Krishna_Mattarella-Micke_Iyyer_2021}
Simeng Sun, Kalpesh Krishna, Andrew Mattarella-Micke, and Mohit Iyyer. 2021.
\newblock Do long-range language models actually use long-range context?
\newblock \emph{arXiv: Computation and Language,arXiv: Computation and Language}.

\bibitem[{Sun et~al.(2022)Sun, Dong, Patra, Ma, Huang, Benhaim, Chaudhary, Song, and Wei}]{sun2022length}
Yutao Sun, Li~Dong, Barun Patra, Shuming Ma, Shaohan Huang, Alon Benhaim, Vishrav Chaudhary, Xia Song, and Furu Wei. 2022.
\newblock A length-extrapolatable transformer.
\newblock \emph{arXiv preprint arXiv:2212.10554}.

\bibitem[{Touvron et~al.(2023{\natexlab{a}})Touvron, Lavril, Izacard, Martinet, Lachaux, Lacroix, Rozi{\`e}re, Goyal, Hambro, Azhar et~al.}]{touvron2023llama1}
Hugo Touvron, Thibaut Lavril, Gautier Izacard, Xavier Martinet, Marie-Anne Lachaux, Timoth{\'e}e Lacroix, Baptiste Rozi{\`e}re, Naman Goyal, Eric Hambro, Faisal Azhar, et~al. 2023{\natexlab{a}}.
\newblock Llama: Open and efficient foundation language models.
\newblock \emph{arXiv preprint arXiv:2302.13971}.

\bibitem[{Touvron et~al.(2023{\natexlab{b}})Touvron, Martin, Stone, Albert, Almahairi, Babaei, Bashlykov, Batra, Bhargava, Bhosale et~al.}]{touvron2023llama}
Hugo Touvron, Louis Martin, Kevin Stone, Peter Albert, Amjad Almahairi, Yasmine Babaei, Nikolay Bashlykov, Soumya Batra, Prajjwal Bhargava, Shruti Bhosale, et~al. 2023{\natexlab{b}}.
\newblock Llama 2: Open foundation and fine-tuned chat models.
\newblock \emph{arXiv preprint arXiv:2307.09288}.

\bibitem[{Tworkowski et~al.(2023)Tworkowski, Staniszewski, Pacek, Wu, Michalewski, and Mi{\l}o{\'s}}]{tworkowski2023focused}
Szymon Tworkowski, Konrad Staniszewski, Miko{\l}aj Pacek, Yuhuai Wu, Henryk Michalewski, and Piotr Mi{\l}o{\'s}. 2023.
\newblock Focused transformer: Contrastive training for context scaling.
\newblock \emph{arXiv preprint arXiv:2307.03170}.

\bibitem[{Vaswani et~al.(2017)Vaswani, Shazeer, Parmar, Uszkoreit, Jones, Gomez, Kaiser, and Polosukhin}]{vaswani2017attention}
Ashish Vaswani, Noam Shazeer, Niki Parmar, Jakob Uszkoreit, Llion Jones, Aidan~N Gomez, {\L}ukasz Kaiser, and Illia Polosukhin. 2017.
\newblock Attention is all you need.
\newblock \emph{Advances in neural information processing systems}, 30.

\bibitem[{Wang et~al.(2020)Wang, Li, Khabsa, Fang, and Ma}]{wang2020linformer}
Sinong Wang, Belinda~Z Li, Madian Khabsa, Han Fang, and Hao Ma. 2020.
\newblock Linformer: Self-attention with linear complexity.
\newblock \emph{arXiv preprint arXiv:2006.04768}.

\bibitem[{Wu et~al.(2022)Wu, Rabe, Hutchins, and Szegedy}]{wu2022memorizing}
Yuhuai Wu, Markus~N Rabe, DeLesley Hutchins, and Christian Szegedy. 2022.
\newblock Memorizing transformers.
\newblock \emph{arXiv preprint arXiv:2203.08913}.

\bibitem[{Xiong et~al.(2023)Xiong, Liu, Molybog, Zhang, Bhargava, Hou, Martin, Rungta, Sankararaman, Oguz et~al.}]{xiong2023effective}
Wenhan Xiong, Jingyu Liu, Igor Molybog, Hejia Zhang, Prajjwal Bhargava, Rui Hou, Louis Martin, Rashi Rungta, Karthik~Abinav Sankararaman, Barlas Oguz, et~al. 2023.
\newblock Effective long-context scaling of foundation models.
\newblock \emph{arXiv preprint arXiv:2309.16039}.

\bibitem[{Ye et~al.(2019)Ye, Guo, Gan, Qiu, and Zhang}]{Ye_Guo_Gan_Qiu_Zhang_2019}
Zihao Ye, Qipeng Guo, Quan Gan, Xipeng Qiu, and Zheng Zhang. 2019.
\newblock Bp-transformer: Modelling long-range context via binary partitioning.
\newblock \emph{arXiv: Computation and Language,arXiv: Computation and Language}.

\bibitem[{Zaheer et~al.(2020)Zaheer, Guruganesh, Dubey, Ainslie, Alberti, Ontanon, Pham, Ravula, Wang, Yang et~al.}]{zaheer2020big}
Manzil Zaheer, Guru Guruganesh, Kumar~Avinava Dubey, Joshua Ainslie, Chris Alberti, Santiago Ontanon, Philip Pham, Anirudh Ravula, Qifan Wang, Li~Yang, et~al. 2020.
\newblock Big bird: Transformers for longer sequences.
\newblock \emph{Advances in neural information processing systems}, 33:17283--17297.

\bibitem[{Zheng et~al.(2023)Zheng, Chiang, Sheng, Zhuang, Wu, Zhuang, Lin, Li, Li, Xing, Zhang, Gonzalez, and Stoica}]{zheng2023judging}
Lianmin Zheng, Wei-Lin Chiang, Ying Sheng, Siyuan Zhuang, Zhanghao Wu, Yonghao Zhuang, Zi~Lin, Zhuohan Li, Dacheng Li, Eric.~P Xing, Hao Zhang, Joseph~E. Gonzalez, and Ion Stoica. 2023.
\newblock \href {http://arxiv.org/abs/2306.05685} {Judging llm-as-a-judge with mt-bench and chatbot arena}.

\end{thebibliography}

\appendix

\section{Training Details}
\label{appdix: training}
The model is trained on 4 NVIDIA A100 GPUs with DeepSpeed~\citep{rasley2020deepspeed}, ZeRO~\citep{rajbhandari2020zero, ren2021zero} Stage 3, gradient-checkpointing~\citep{chen2016training}, and FlashAttention~\citep{dao2022flashattention, dao2023flashattention}. We also use BF16 and TF32 mix computation precision for further acceleration. 

All the models are fine-tuned using AdamW Optimizer~\citep{Loshchilov_Hutter_2017} with $\beta_1=0.9$ and $\beta_2=0.95$ for two epochs, computing losses on response tokens only. We use a cosine learning rate scheduler, set the peak learning rate to 2e-5, and weight decay to 0.1. For training on 3.5k conversations, we use a batch size of 128 and 10 warmup steps. We use a batch size of 32 and 0 warmup steps for fewer training data. If not explicitly stated, we default to using 3.5k ShareGPT conversations for instruction tuning.

\section{Additional Experiment Results}
\label{appdix: exp}

\begin{table*}[!t]
    \centering
    \resizebox{\textwidth}{!}{
        \begin{tabular}{lccccccccccccc}
        \toprule
        \multirow{2}{*}{\textbf{\Large Model}} & \multicolumn{3}{c}{\textbf{\Large Singl-Doc QA}} & \multicolumn{3}{c}{\textbf{\Large Multi-Doc QA}} & \multicolumn{3}{c}{\textbf{\Large Summarization}} & \multicolumn{3}{c}{\textbf{\Large Few-shot Learning}} & \multirow{2}{*}{\textbf{\Large Macro}} \\
        \cmidrule(lr){2-4} \cmidrule(lr){5-7} \cmidrule(lr){8-10} \cmidrule(lr){11-13} 
        & \textbf{\Large NQA} & \textbf{\Large QAPR} & \textbf{\Large MFQA\_en} & \textbf{\Large HPQA} & \textbf{\Large WMQA} & \textbf{\Large MSQ} & \textbf{\Large GR} & \textbf{\Large QMSM} & \textbf{\Large MNWS} & \textbf{\Large TREC} & \textbf{\Large TRVQA} & \textbf{\Large SMSM} \\ \midrule
        PI-4K & 22.3 & 27.3 & 44.6 & 31.7 & 30.8 & 9.2 & 29.3 & 21.5 & 27.1 & 61.5 & 87.5 & 41.4 & 36.2 \\
        PI-8K & 21.4 & 28.6 & 46.8 & 31.1 & 29 & 11.7 & 30.1 & 22.5 & 27.1 & 66 & 86.8 & 42.2 & 36.9 \\
        PI-16K & 20.1 & 30.4 & 45.3 & 26.1 & 30.1 & 9.9 & 28.1 & 23.7 & 26.6 & 68 & 84.9 & 42.5 & 36.3 \\
        PI-32K & 7 & 27.6 & 43.7 & 16.3 & 25.1 & 1.5 & 23.5 & 14.2 & 27 & 67.5 & 54.4 & 33.1 & 28.4 \\
        NTK-By-Parts-4k & 22.8 & 27.3 & 42.5 & 26.5 & 23 & 10.1 & 28.7 & 21.8 & 27.1 & 63 & 87.3 & 42 & 35.2 \\
        NTK-By-Parts-8k & 20.9 & 31.2 & 43.5 & 28.3 & 29.4 & 10.9 & 29.4 & 22.3 & 27 & 65 & 87.5 & 43.5 & 36.6 \\
        NTK-By-Parts-16k & 15.9 & 31.1 & 40.1 & 25.4 & 26.6 & 7.2 & 26.7 & 22.4 & 26.9 & 68.5 & 82.8 & 42.9 & 34.7 \\
        NTK-By-Parts-32k & 6.5 & 30.8 & 39.1 & 15.9 & 26.2 & 1.1 & 23.3 & 14.9 & 26.9 & 68.5 & 54.4 & 34.7 & 28.5 \\
        Yarn-4K & 21 & 27.9 & 43.8 & 29 & 29.5 & 12.8 & 29.3 & 22.1 & 26.9 & 61.5 & 87.1 & 42.8 & 36.1 \\
        Yarn-8K & 20.9 & 31.2 & 43.5 & 28.3 & 29.4 & 10.9 & 29.4 & 22.3 & 27 & 65 & 87.5 & 43.5 & 36.6 \\
        Yarn-16K & 20.3 & 28.9 & 42.8 & 27.8 & 30.7 & 7.2 & 27.4 & 22.5 & 26.8 & 66 & 85.6 & 42.6 & 35.7 \\
        Yarn-32K & 6.5 & 29.3 & 39 & 16 & 29.1 & 1.3 & 24 & 14.4 & 26.9 & 66.5 & 55.1 & 34.6 & 28.6 \\
        ABF-4K & 19.5 & 30.6 & 44.1 & 31.1 & 29.5 & 11.5 & 29 & 21.4 & 27.6 & 64.5 & 87.6 & 41.9 & 36.5 \\
        ABF-8K & 22.7 & 32.1 & 45.2 & 35.5 & 29.2 & 14.8 & 30.6 & 22.8 & 27.4 & 67 & 84.8 & 42.4 & 37.9 \\
        ABF-16K & 24.6 & 32.8 & 45.6 & 35.1 & 30.3 & 15.2 & 30.8 & 23 & 27.4 & 71 & 84.7 & 42.7 & 38.6 \\
        ABF-32K & 23.6 & 27.1 & 44.8 & 36.8 & 27.5 & 12.3 & 29.4 & 23.1 & 26.5 & 72 & 84.9 & 43.5 & 37.6 \\
        Ours-4K & 21.1 & 30.1 & 43.3 & 28.1 & 31.6 & 11.1 & 28.7 & 21.6 & 27.6 & 64 & 86.8 & 42.1 & 36.3 \\
        Ours-8K & 23.3 & 31.7 & 45.5 & 33.2 & 31.7 & 14.6 & 30.7 & 23 & 27 & 67.5 & 86.3 & 42.4 & 38.1 \\
        Ours-16K & 21.9 & 31 & 47.1 & 40.1 & 32.7 & 15.1 & 32.3 & 23 & 27.1 & 70.5 & 86.7 & 42 & 39.1 \\
        Ours-32K & 23.6 & 31.8 & 45.5 & 39 & 31.7 & 16.6 & 31.7 & 23.6 & 27.1 & 70.5 & 86 & 42.4 & 39.1 \\ 
        \bottomrule
        \end{tabular}
    }
    \vspace{3pt}
    \caption{Long-Context performance of RoPE-extension Methods with different context window sizes}
    \label{tab: ctx}
\end{table*}

\begin{table*}[!t]
    \centering
    \resizebox{\textwidth}{!}{
        \begin{tabular}{lccccccccccccc}
        \toprule
        \multirow{2}{*}{\textbf{\Large Model}} & \multicolumn{3}{c}{\textbf{\Large Singl-Doc QA}} & \multicolumn{3}{c}{\textbf{\Large Multi-Doc QA}} & \multicolumn{3}{c}{\textbf{\Large Summarization}} & \multicolumn{3}{c}{\textbf{\Large Few-shot Learning}} & \multirow{2}{*}{\textbf{\Large Macro}} \\
        \cmidrule(lr){2-4} \cmidrule(lr){5-7} \cmidrule(lr){8-10} \cmidrule(lr){11-13} 
        & \textbf{\Large NQA} & \textbf{\Large QAPR} & \textbf{\Large MFQA\_en} & \textbf{\Large HPQA} & \textbf{\Large WMQA} & \textbf{\Large MSQ} & \textbf{\Large GR} & \textbf{\Large QMSM} & \textbf{\Large MNWS} & \textbf{\Large TREC} & \textbf{\Large TRVQA} & \textbf{\Large SMSM} \\ \midrule
        PI-32 & 10.7 & 15.2 & 30.5 & 18.5 & 21.6 & 7.3 & 31 & 21.2 & 27.5 & 60 & 73.2 & 39.4 & 29.7 \\
        PI-100 & 7.2 & 31.4 & 37.1 & 30.9 & 33.4 & 11.6 & 30.5 & 16 & 27.1 & 64.5 & 78.4 & 38 & 33.8 \\
        PI-1000 & 20.4 & 31.1 & 39 & 30.7 & 30.5 & 10.4 & 27.9 & 23.5 & 26.6 & 67 & 84.5 & 41.6 & 36.1 \\
        PI-3500 & 20.1 & 30.4 & 45.3 & 26.1 & 30.1 & 9.9 & 28.1 & 23.7 & 26.6 & 68 & 84.9 & 42.5 & 36.3 \\
        NTK-By-Parts-32 & 4.5 & 22.6 & 31.9 & 10.9 & 23.7 & 0.8 & 20.7 & 13.2 & 26.3 & 65 & 41.3 & 33.4 & 24.5 \\
        NTK-By-Parts-100 & 7.9 & 28.6 & 40.1 & 19.1 & 26.9 & 7 & 24.7 & 18 & 26.1 & 66.5 & 77.3 & 40 & 31.9 \\
        NTK-By-Parts-1000 & 15.9 & 28.3 & 42.7 & 23.6 & 26.5 & 6.5 & 26.4 & 22.7 & 26.7 & 68.5 & 83.7 & 42.1 & 34.5 \\
        NTK-By-Parts-3500 & 15.9 & 31.1 & 40.1 & 25.4 & 26.6 & 7.2 & 26.7 & 22.4 & 26.9 & 68.5 & 82.8 & 42.9 & 34.7 \\
        Yarn-32 & 5 & 18.9 & 35.1 & 12 & 26 & 0.9 & 23.6 & 14.1 & 26.2 & 63.5 & 44.1 & 30.4 & 25 \\
        Yarn-100 & 6.6 & 30.1 & 39.2 & 17.7 & 27.1 & 2.8 & 24.4 & 16.7 & 25.8 & 66.5 & 76 & 37.9 & 30.9 \\
        Yarn-1000 & 18 & 28.2 & 42.9 & 26.7 & 28.4 & 9.3 & 27.6 & 22.4 & 26.9 & 65 & 85.1 & 41.9 & 35.2 \\
        Yarn-3500 & 19.7 & 25.4 & 44.6 & 29.3 & 25.9 & 9.5 & 26.5 & 22.2 & 26.7 & 67.5 & 85.6 & 43.7 & 35.5 \\
        ABF-32 & 10.9 & 15 & 32.2 & 20.3 & 21.6 & 7.5 & 28.3 & 21.2 & 26.9 & 56 & 41.6 & 35.2 & 26.4 \\
        ABF-100 & 18.8 & 27.6 & 41.2 & 30.9 & 35 & 10.2 & 31.3 & 22 & 27.2 & 66.5 & 73.8 & 36.7 & 35.1 \\
        ABF-1000 & 23.9 & 33 & 44.9 & 32 & 20.1 & 12.2 & 31.1 & 23.9 & 27.5 & 71 & 85.6 & 40.7 & 37.2 \\
        ABF-3500 & 24 & 30.5 & 45.8 & 37.9 & 30.7 & 15.4 & 31.4 & 23.3 & 27.2 & 70 & 84.2 & 42.6 & 38.6 \\
        Ours-32 & 14.8 & 15.6 & 36.4 & 29.6 & 25.9 & 12.9 & 32.2 & 21.4 & 26.9 & 55 & 67.3 & 38.1 & 31.3 \\
        Ours-100 & 20.6 & 26.4 & 45.9 & 37.7 & 35.6 & 16.4 & 32.2 & 22 & 26.9 & 67.5 & 80.2 & 37.3 & 37.4 \\
        Ours-1000 & 23.5 & 33 & 45 & 34.3 & 24.8 & 16.4 & 30.9 & 23.8 & 27.9 & 71 & 87.7 & 41 & 38.3 \\
        Ours-3500 & 21.9 & 31 & 47.1 & 40.1 & 32.7 & 15.1 & 32.3 & 23 & 27.1 & 70.5 & 86.7 & 42 & 39.1 \\ 
        \bottomrule
        \end{tabular}
    }
    \vspace{3pt}
    \caption{Long-context performance of RoPE-extension methods with different amounts of training data}
    \label{tab: data}
\end{table*}

\end{document}